\documentclass{article}

\usepackage{arxiv}

\usepackage[utf8]{inputenc} 
\usepackage[T1]{fontenc}    
\usepackage{hyperref}       
\usepackage{url}            
\usepackage{booktabs}       
\usepackage{amsfonts}       
\usepackage{nicefrac}       
\usepackage{microtype}      
\usepackage{lipsum}
\usepackage{graphicx}
\usepackage{xcolor}
\usepackage{multirow}
\graphicspath{ {./images/} }

\title{\textbf{LLaSM: \linebreak Large Language and Speech Model}}

\begin{document}
\maketitle
\begin{center}
\large
    Yu Shu\textsuperscript{1*}, Siwei Dong\textsuperscript{1*}, Guangyao Chen\textsuperscript{1,2*}, Wenhao Huang\textsuperscript{3},
    \linebreak Ruihua Zhang, Daochen Shi, Qiqi Xiang \& Yemin Shi\textsuperscript{1$\dag$}
    
\small
   \textsuperscript{1}LinkSoul.AI, \textsuperscript{2}Peking University, \textsuperscript{3}01.ai

\small
   \textsuperscript{*} Equal contribution \linebreak
   \textsuperscript{$\dag$} Corresponding author: ymshi@linksoul.ai

\end{center}
~
~
\vspace{8pt}

\begin{abstract}
Multi-modal large language models have garnered significant interest recently. 
Though, most of the works focus on vision-language multi-modal models providing strong capabilities in following vision-and-language instructions. 
However, we claim that speech is also an important modality through which humans interact with the world. 
Hence, it is crucial for a general-purpose assistant to be able to follow multi-modal speech-and-language instructions. 
In this work, we propose \textbf{L}arge \textbf{L}anguage \textbf{a}nd \textbf{S}peech \textbf{M}odel (\textbf{LLaSM}).
LLaSM is an end-to-end trained large multi-modal speech-language model with cross-modal conversational abilities, capable of following speech-and-language instructions. 
Our early experiments show that LLaSM demonstrates a more convenient and natural way for humans to interact with artificial intelligence. 
Specifically, we also release a large Speech Instruction Following dataset LLaSM-Audio-Instructions. 
Code and demo are available at \url{https://github.com/LinkSoul-AI/LLaSM} and \url{https://huggingface.co/spaces/LinkSoul/LLaSM}. The LLaSM-Audio-Instructions dataset is available at \url{https://huggingface.co/datasets/LinkSoul/LLaSM-Audio-Instructions}.
\end{abstract}


\section{Introduction}
Speech contains semantic information and contains paralinguistic information like intonation at the same time, it carries more quantity of information than text. Additionally, speech is a more convenient and natural way for humans to interact with artificial intelligence. Therefore, following speech-and-language instructions is crucial when developing a general-purpose assistant.

However, most large language models \cite{brown2020language,chowdhery2022palm,touvron2023llama} receive text input only, which restricts the ability of large language models. Vision-and-language multi-modal models \cite{openai2023gpt4,driess2023palme,liu2023llava,zhu2023minigpt4,girdhar2023imagebind,Emu} offer the ability to understand the vision information, making a huge step toward general artificial intelligence (AGI), but it is still inconvenient for humans to input the tasks by typing a text instruction. The cascading paradigm methods \cite{shen2023hugginggpt,huang2023audiogpt} use an automatic speech recognition (ASR) model to convert the speech input into the text input, then the model can process the task with the text input. However, it still leads to information consumption during the modal transformation from speech to text and might import mistakes of the ASR system. Recently, speech-language multi-modal models \cite{zhang2023speechgpt,rubenstein2023audiopalm} focusing on processing and generating speech and text with a large language model are capable of understanding and generating multi-modal content. The speech signals are encoded into discrete tokens, and then discrete speech tokens are expanded into the vocabulary of the LLM. In this way, the LLM needs to be retrained with plenty of multi-modal data and huge computing resources.

In this paper, we propose LLaSM, a large speech-and-language model with cross-modal conversational abilities, capable of understanding and following speech-and-language instructions. Following the manner of LLaVA \cite{liu2023llava}, we leverage the well-trained speech modal encoder and the LLM, which makes LLaSM more resource-friendly.
Specifically, we use Whisper \cite{radford2022whisper} as a speech encoder to encode the speech signals into embeddings. Then a modal adaptor learns to align speech embeddings with the input text embeddings of the large language model. The speech embeddings and the text embeddings are concatenated together to form interleaved sequences, then the interleaved sequences are input to the LLM for supervised fine-tuning.
The training process is divided into two stages.
In the first stage, we use the public ASR datasets for the modality adaptation pre-training. The speech encoder and the LLM are frozen, only the modal adaptor is trained to align the speech and text embeddings. As most of the model parameters remain frozen, only a small part of the parameters from the modal adaptor is trained during this stage, it is not resource-consuming.
In the second stage, we use cross-modal instruction data for training to provide the model with the capacity to process cross-modal conversations and handle multi-modal instructions.
The speech encoder is frozen while the parameters of the modal adaptor and the language model are updated for cross-modal instruction fine-tuning.
Worth noting that existing open-source speech-text cross-modal instruction-following datasets are scarce, so we build and release a speech-text cross-modal instruction-following dataset \textbf{LLaSM-Audio-Instructions}. The dataset is constructed by carefully selecting dialogues from GPT4-LLM \cite{peng2023instruction}, ShareGPT \cite{sharegpt}, WizardLM \cite{xu2023wizardlm}, and using text-to-speech technology to generate a large amount of dialogue audio data. In total, it contains 199k conversations, in which there are 80k Chinese audio samples and 428k English audio samples, which is the largest Chinese and English speech-text cross-modal instruction-following dataset to our knowledge.

Our paper makes the following contributions:

\begin{itemize}
    \item We build a speech-language multi-modal model that can understand and follow speech-language instructions, which provides a more convenient and natural way for humans to interact with artificial intelligence.
    \item We construct and release LLaSM-Audio-Instrustions, a large-scale Chinese and English speech-text cross-modal instruction-following dataset. We release the data in \url{https://huggingface.co/datasets/LinkSoul/LLaSM-Audio-Instructions}.
    \item We release the code in \url{https://github.com/LinkSoul-AI/LLaSM} and the demo is shown in \url{https://huggingface.co/spaces/LinkSoul/LLaSM}.
    
\end{itemize}

\section{Related Work}
\textbf{Vision Large Language Model} has gained significant traction \cite{openai2023gpt4,driess2023palme,liu2023llava,zhu2023minigpt4,girdhar2023imagebind,Emu} recently. Most of them leverage the pre-trained LLMs and vision encoders to perform vision tasks. Flamingo \cite{alayrac2022flamingo} aligns a pre-trained vision encoder and language model using gated cross-attention and is trained on billions of image-text pairs. BLIP-2 \cite{li2023blip2} employs a Flan-T5 \cite{chung2022scaling} with a Q-Former to efficiently align visual features with the language model. Palm-E \cite{driess2023palme}, featuring 562 billion parameters, integrates the 540B PaLM \cite{chowdhery2022palm} and 22B Vision Transformer \cite{dosovitskiy2021image} into the largest vision-language model. LLaVA \cite{liu2023llava} leverages pre-trained CLIP \cite{radford2021learning} visual encoder and LLaMA \cite{touvron2023llama} and conducts instruct tuning on GPT4-assisted visual instruction data. GPT-4 \cite{openai2023gpt4} also shows powerful visual understanding and reasoning abilities. The success of the multi-modal large language model in the visual domains has brought a lot of inspiration to the research in the speech domains as well.

\textbf{Speech Large Language Model} has gained more and more interest, for the success of the vision multi-modal LLMs. The cascading paradigm methods \cite{shen2023hugginggpt,huang2023audiogpt} use an automatic speech recognition (ASR) model to convert the speech input into the text input, which still leads to information consumption and might import mistakes of the ASR system. Recently, speech-language multi-modal models \cite{zhang2023speechgpt,rubenstein2023audiopalm} focusing on processing and generating speech and text with a large language model are capable of understanding and generating multi-modal content. The speech signals are encoded into discrete tokens, and then discrete speech tokens are expanded into the vocabulary of the LLM. In this way, the LLM needs to be retrained with plenty of multi-modal data and huge computing resources.

\section{Approach}

\subsection{Model}
\begin{figure}
    \centering
    \includegraphics[width=1.0\textwidth]{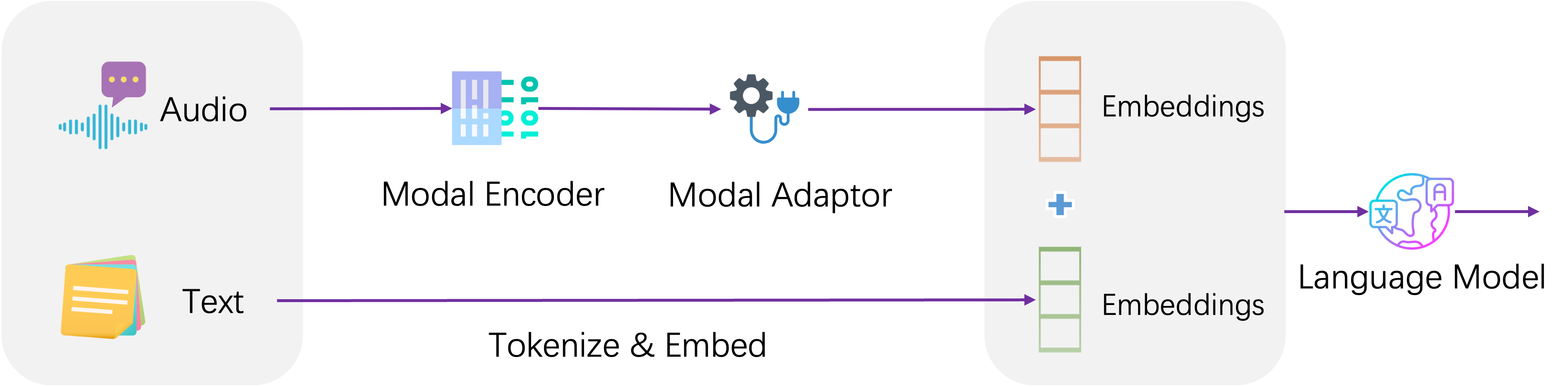}
    \caption{Model framework of the LLaSM}
    \label{fig:model_framework}
\end{figure}

The focus of training multi-modal models is to fuse cross-modal complementary information of multi-modalities and effectively exploit the capabilities of well-trained large language models. The LLaSM model architecture is shown in Figure \ref{fig:model_framework}. We use Whisper \cite{radford2022whisper} to encode the raw audio data into embeddings first, then a modal adaptor is trained during the pre-training stage to align the audio embeddings and the text embeddings. The audio embeddings and the text embeddings are concatenated together to form interleaved input sequences to input to the large language model. We choose Chinese-LLAMA2-7B \cite{Shu2023ChineseLlama2} as our LLM, for its capabilities in both Chinese and English. During the cross-modal instruction fine-tuning stage, the modal adaptor and the LLM are trained with multi-tasks.

\begin{figure}[htbp]
\centering
\begin{minipage}[t]{0.48\textwidth}
\centering
\includegraphics[scale=0.40]{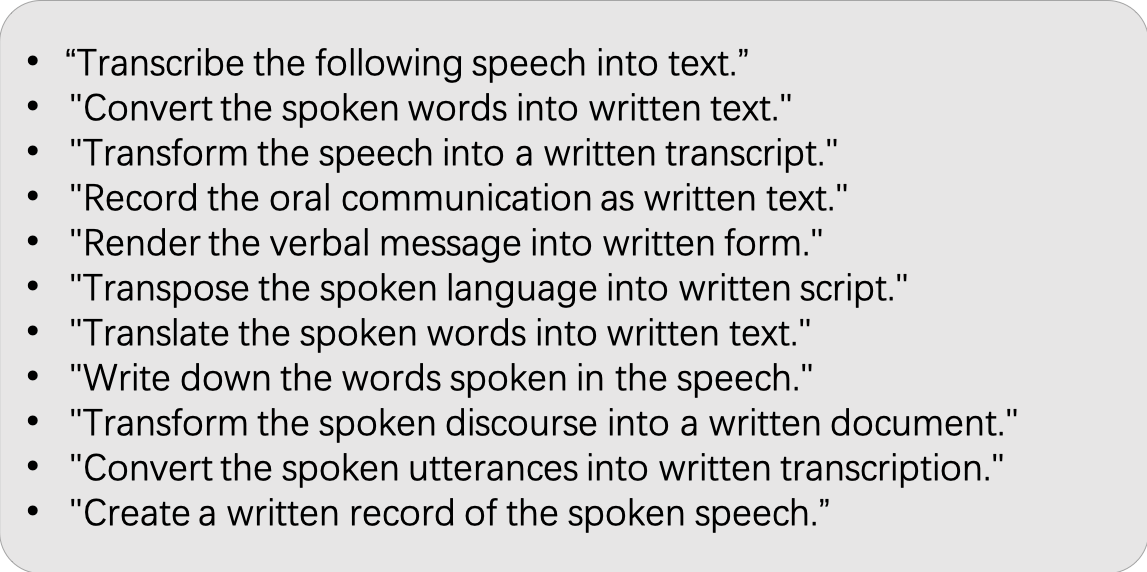}
\caption{English simple instructions.}
\label{fig:en_prompt}
\end{minipage}
\begin{minipage}[t]{0.48\textwidth}
\centering
\includegraphics[scale=0.40]{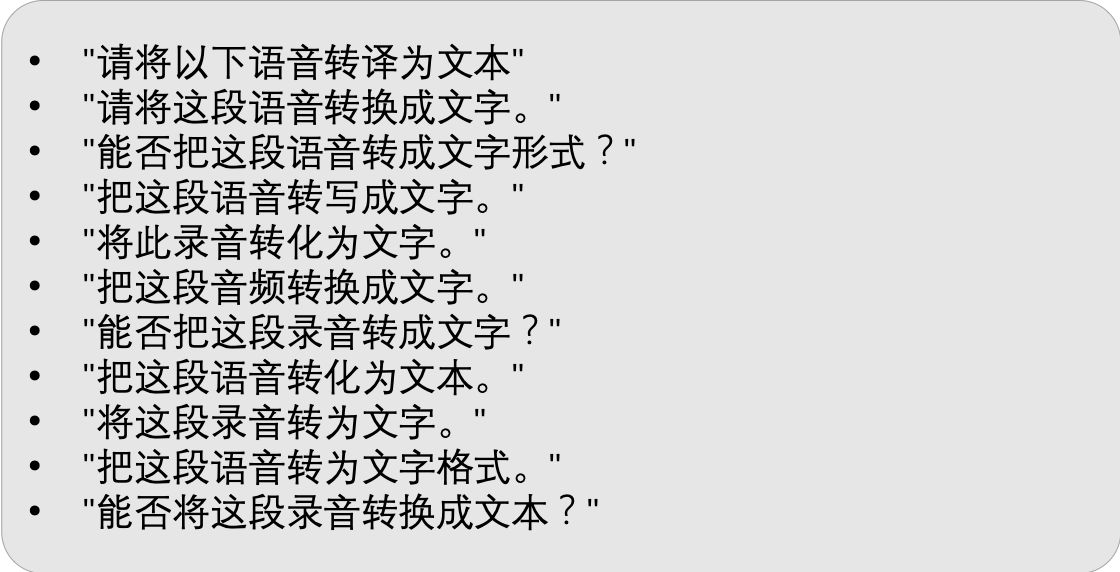}
\caption{Chinese simple instructions.}
\label{fig:zh_prompt}
\end{minipage}
\end{figure}

\begin{figure}
    \centering
    \includegraphics[width=1.0\textwidth]{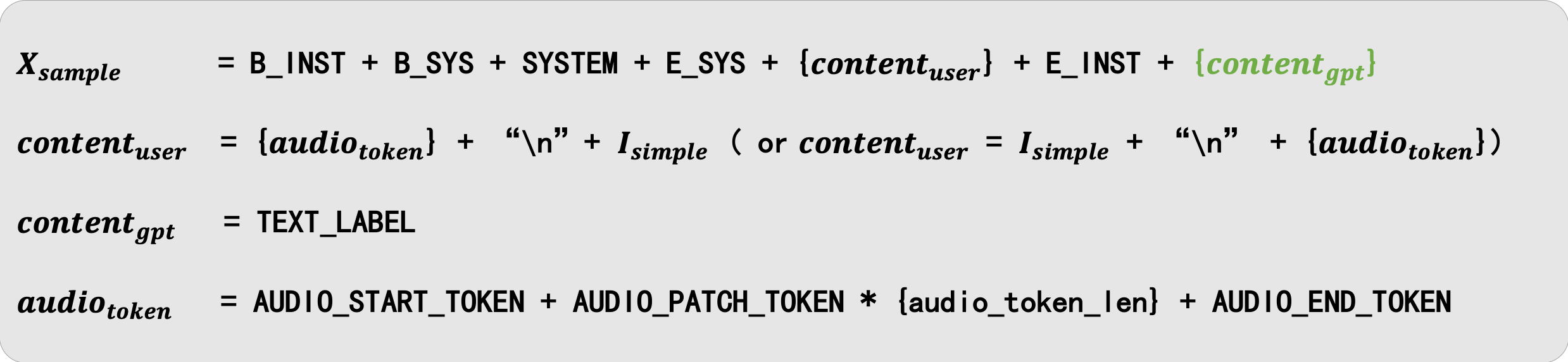}
    \caption{The sample sequence format for the pre-training. We follow the manner of Llama-2, and B\_INST = '[INST]', E\_INST = '[/INST]', B\_SYS = '$<<$SYS$>>\backslash$n', E\_SYS = '$\backslash n<<$/SYS$>>\backslash$n$\backslash$n'. The SYSTEM = 'You are a helpful language and speech assistant. You are able to understand the speech content that the user provides, and assist the user with a variety of tasks using natural language.', and the TEXT\_LABEL is the text label of the ASR data sample. The audio\_token\_len is set to 64 by default. Special audio tokens are used, AUDIO\_START\_TOKEN = '<au\_start>', AUDIO\_END\_TOKEN = '<au\_end>', AUDIO\_PATCH\_TOKEN = '<au\_patch>'. The $content_{user}$ consists of the $audio_{token}$ and the $I_{simple}$, in which $I_{simple}$ is a simple instruction and is randomly put before or after the $audio_{token}$. While training the BOS token and the EOS token will be added to each sample at the beginning and the end of the sequence, only the {\color{green}green tokens} are used to compute the loss.}
    \label{fig:simple_template}
\end{figure}

\textbf{The pre-training stage.} During this stage, the modal encoder and the LLM remain frozen. To enable the LLM to understand the audio embeddings from the modal encoder, the modal adaptor is trained with public ASR data to align the text and the audio embeddings. The data sample (audio data, text label) of ASR data is formatted as a tuple of (simple instruction, audio data, text label), in which the simple instruction is an automatic speech recognition instruction. According to the different languages of the audio data, an English simple instruction listed in Figure \ref{fig:en_prompt} or a Chinese simple instruction listed in Figure \ref{fig:zh_prompt} will be chosen. The unified format of the pre-training multi-modal sequence $X_{sample}$ is shown in Figure \ref{fig:simple_template}. Each data sample is formatted as $X_{sample}$, then we will replace the audio patch embeddings from the text sequence with the audio embeddings of the modal adaptor. The final interleaved input embeddings will be input to the large language model. The training target is to predict the text label of each data sample.

\begin{figure}
    \centering
    \includegraphics[width=1.0\textwidth]{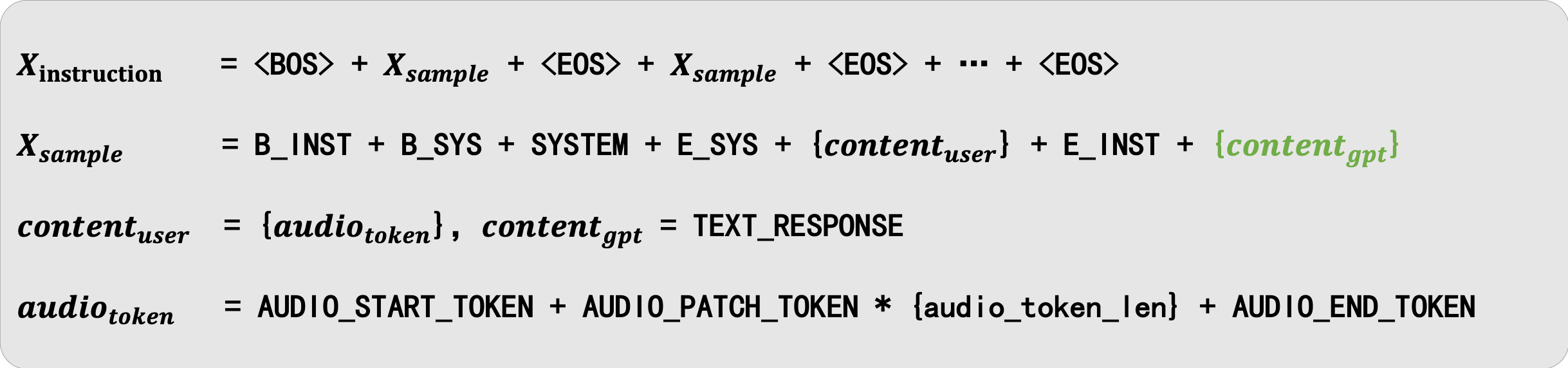}
    \caption{The sample sequence format for the cross-modal instruction fine-tuning. We follow the manner of Llama-2, and B\_INST = '[INST]', E\_INST = '[/INST]', B\_SYS = '$<<$SYS$>>\backslash$n', E\_SYS = '$\backslash n<<$/SYS$>>\backslash$n$\backslash$n'. The SYSTEM = 'You are a helpful language and speech assistant. You are able to understand the speech content that the user provides, and assist the user with a variety of tasks using natural language.', and the TEXT\_RESPONSE is the text response from the chatbot. The audio\_token\_len is set to 64 by default. Special audio tokens are used, AUDIO\_START\_TOKEN = '<au\_start>', AUDIO\_END\_TOKEN = '<au\_end>', AUDIO\_PATCH\_TOKEN = '<au\_patch>'. The $content_{user}$ is the $audio_{token}$ which will be replaced by the audio embeddings during training. Each round of question and answer will be formatted as $X_{sample}$, which will be concatenated together with the EOS token. While training the BOS token will be added at the beginning of the sequence, and the EOS token will be added at the end of the sequence, only the {\color{green}green tokens} are used to compute the loss.}
    \label{fig:instruct_template}
\end{figure}

\textbf{The cross-modal instruction fine-tuning.} During this stage, only the modal encoder is frozen, the modal adaptor and the LLM are joint-trained with multi-tasks. We build complex cross-modal instructions using several conversational data. The questions from humans are generated to audio data by using Microsoft Azure text-to-speech API, then the training target is to predict the responses from the chatbot. A round of question and answer will be processed into a multi-modal sequence $X_{sample}$, and multiple rounds of question and answer will be concatenated with the EOS token. The unified format of the cross-modal instruction fine-tuning sequence is shown in Figure \ref{fig:instruct_template}. As the effectiveness of text-only conversational data with multi-task instructions has been demonstrated in several open-source language-only instruction-tuning works \cite{peng2023instruction,sharegpt,xu2023wizardlm}, the cross-modal instructions are able to improve the capacity of following multi-modal instructions. 

\subsection{Data Collection}
To enable the LLM to understand the audio signals, we collect several public ASR data sets to form the Modality Adaptation Pre-training Data with simple instructions of automatic speech recognition. And, for cross-modal instruction tuning, we use several open-source language-only instruction-tuning data sets to build the Cross-modal Instruction Fine-Tuning Data by generating the speech data. The details are as follows.

\textbf{Modality Adaptation Pre-training Data.}
To align the embeddings of text and audio, we collect several public ASR data sets in both English and Chinese, including Aishell \cite{aishell_2017}, LibriSpeech \cite{panayotov2015librispeech}, Magicdata \cite{magicdata} and Primewords \cite{primewords_201801}. The data sample of ASR data usually consists of a pair of speech audio and text utterances, especially, when we add a simple instruction to the data sample as the task instruction. These simple instructions are listed in Figure \ref{fig:en_prompt} and Figure \ref{fig:zh_prompt}, which are different representations of the automatic speech recognition task in both English and Chinese. While pre-training, the simple instruction and the audio data are input to the model to predict the text label of the audio data.

\begin{figure}
    \centering
    \includegraphics[width=1.0\textwidth]{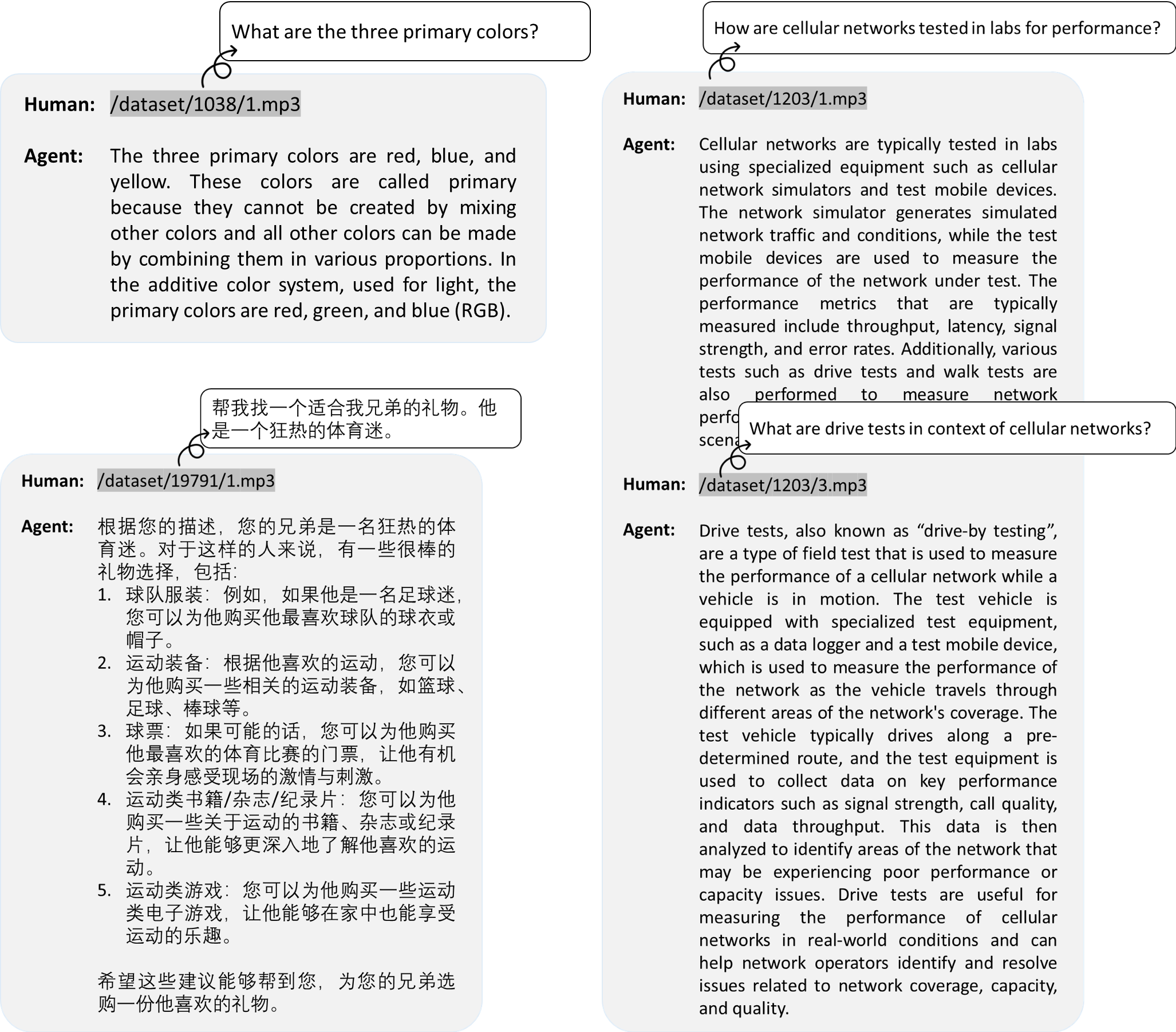}
    \caption{Data samples of the LLaSM-Audio-Instructions.}
    \label{fig:data_samples}
\end{figure}

\textbf{Cross-modal Instruction Fine-Tuning Data.}
As the effectiveness of the open-source language-only instruction-tuning data sets has been demonstrated in previous works\cite{peng2023instruction,sharegpt,xu2023wizardlm}, a natural idea is to generate audio data of these language-only data sets to build a cross-modal instruction-tuning data.
In the process of building this dataset, we first carefully filtered all the conversation data, by removing the conversations that are not suitable for vocalization, including codes, a large number of symbols, URLs, and other non-readable text. To ensure the data quality, in the second stage, all the answers from chat-bots in the conversations are filtered again. Those that do not contain valuable information are dropped. In the third stage, we use Microsoft Azure text-to-speech API \cite{MicrosoftAzure} to generate speech data from humans in these data sets. The speech data of humans are used as the complex instructions and the responses from the chatbot are predicted during the instruction fine-tuning.
Specifically, 80k conversation data which contains 160k samples is selected from WizardLM \cite{xu2023wizardlm}, 23k conversation data which contains 155k samples is selected from ShareGPT \cite{sharegpt} and 96k conversation data which contains 192k samples is selected from GPT-4-LLM \cite{peng2023instruction}. Table \ref{tab:data} shows the specific details of the dataset, which contains 199k conversation data and 508k samples in total.
Several examples of the dataset are shown in Figure \ref{fig:data_samples}.
We release this dataset as LLaSM-Audio-Instructions at \url{https://huggingface.co/datasets/LinkSoul/LLaSM-Audio-Instructions}.

\begin{table}
 \caption{LLaSM-Audio-Instructions Data.}
  \centering
  \begin{tabular}{ccccc}
    \toprule
    \multicolumn{5}{c}{LLaSM-Audio-Instructions} \\
    \midrule
    Source     & Conversations & Samples & English Samples     & Chinese Samples \\
    \midrule

     WizardLM & 80k & 160k & 159k  & <1k     \\
     ShareGPT  & 23k  & 155k & 140k & 15k      \\
     GPT-4-LLM & 96k & 192k  & 128k  & 64k  \\
     \midrule
     Total & 199k & 508k & 428k & 80k \\
    \bottomrule
  \end{tabular}
  \label{tab:data}
\end{table}

\begin{figure}
    \centering
    \includegraphics[width=1.0\textwidth]{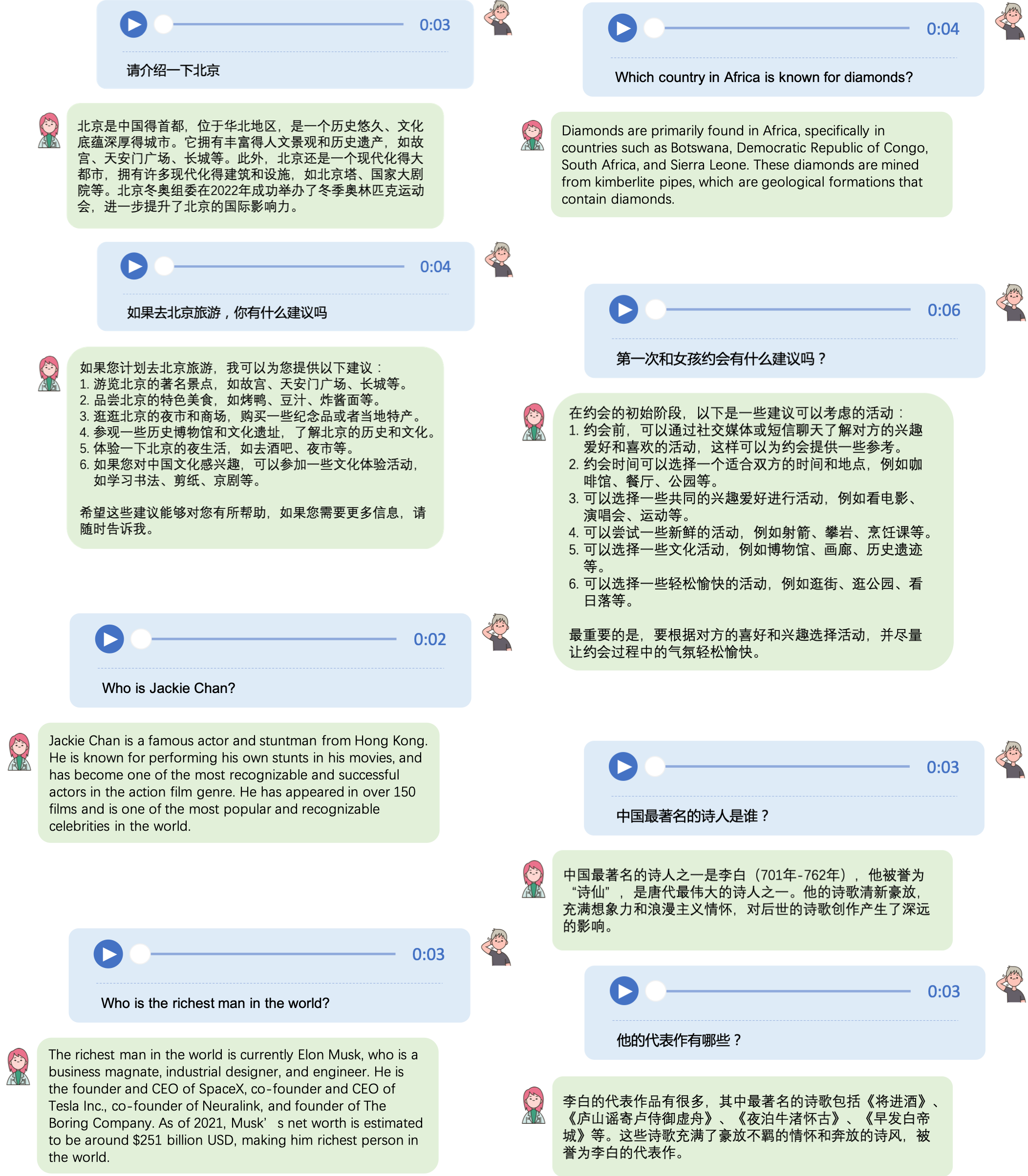}
    \caption{Examples of experiments.}
    \label{fig:experiments}
\end{figure}


\section{Experiments}
As shown in Figure \ref{fig:experiments}, our proposed model, LLaSM, can adaptively recognize and respond to speech in Chinese and English. Figure \ref{fig:experiments} further demonstrates the effectiveness of LLaSM in a bilingual setting. Unlike conventional models that rely on speech-to-text conversion as a preprocessing step, LLaSM can directly process speech inputs, which improves its execution efficiency. Furthermore, LLaSM can support multiple languages and scenarios, which expands its application range. Therefore, LLaSM is a promising model for convenient and interactive human-artificial intelligence communication.




\section{Conclusion}
This work presents LLaSM, a large language model with cross-modal conversational abilities, capable of understanding and following speech-and-language instructions. Experiments show that LLaSM demonstrates a more convenient and natural way for humans to interact with artificial intelligence. Specifically, to alleviate the scarcity of cross-modal speech-and-language instructions data, we build a large Speech Instruction Following data set LLaSM-Audio-Instructions. It is the largest Chinese and English speech-text cross-modal instruction-following data set to our knowledge. Finally, by adopting a visual modal encoder that can easily provide LLaSM with visual capabilities, we will explore combining both vision and audio modalities in future work.

\bibliographystyle{unsrt}  
\bibliography{references}  






\end{document}